\documentclass[letterpaper]{article} 
\usepackage{aaai2026}  
\usepackage{times}  
\usepackage{helvet}  
\usepackage{courier}  
\usepackage[hyphens]{url}  
\usepackage{graphicx} 
\usepackage{natbib}  
\usepackage{caption} 
\usepackage{algorithm}
\usepackage{algorithmic}

\usepackage{amsmath}
\usepackage{cleveref}
\usepackage{amssymb}
\usepackage{adjustbox}
\usepackage{booktabs}
\usepackage{pifont}

\crefname{figure}{Fig.}{Figs.}
\crefname{table}{Tab.}{Tabs.}
\crefname{section}{Sec.}{Secs.}
\crefname{subsection}{Sec.}{Secs.}
\crefname{equation}{Eq.}{Eqs.}
\crefname{appendix}{Appendix}{Appendices}

\usepackage{newfloat}
\usepackage{listings}
\DeclareCaptionStyle{ruled}{labelfont=normalfont,labelsep=colon,strut=off} 
\floatstyle{ruled}
\newfloat{listing}{tb}{lst}{}
\floatname{listing}{Listing}
\title{Human-Centric Open-Future Task Discovery: Formulation, Benchmark, \\and Scalable Tree-Based Search}
\author{
    Zijian Song\textsuperscript{\rm 1},
    Xiaoxin Lin\textsuperscript{\rm 1},
    Tao Pu\textsuperscript{\rm 1},
    Zhenlong Yuan\textsuperscript{\rm 4},
    Guangrun Wang\textsuperscript{\rm 1,\rm 2,\rm 3}\thanks{corresponding author},
    Liang Lin\textsuperscript{\rm 1,\rm 2,\rm 3}
}
\affiliations{
    \textsuperscript{\rm 1}Sun Yat-sen University\\
    \textsuperscript{\rm 2}Guangdong Key Laboratory of Big Data Analysis and Processing\\
    \textsuperscript{\rm 3}X-Era AI Lab\\
    \textsuperscript{\rm 4}AMAP, Alibaba

    \{songzj8,linxx76\}@mail2.sysu.edu.cn,
    putao537@gmail.com,
    yuanzhenlong.yzl@alibaba-inc.com\\
    wanggrun@gmail.com,
    linliang@ieee.org
}

\usepackage{bibentry}

\begin{document}

\maketitle

\begin{abstract}
Recent progress in robotics and embodied AI is largely driven by Large Multimodal Models (LMMs). However, a key challenge remains underexplored: how can we advance LMMs to discover tasks that assist humans in open-future scenarios, where human intentions are highly concurrent and dynamic. In this work, we formalize the problem of Human-centric Open-future Task Discovery (HOTD), focusing particularly on identifying tasks that reduce human effort across plausible futures. To facilitate this study, we propose HOTD-Bench, which features over 2K real-world videos, a semi-automated annotation pipeline, and a simulation-based protocol tailored for open-set future evaluation. Additionally, we propose the Collaborative Multi-Agent Search Tree (CMAST) framework, which decomposes complex reasoning through a multi-agent system and structures the reasoning process through a scalable search tree module. In our experiments, CMAST achieves the best performance on the HOTD-Bench, significantly surpassing existing LMMs. It also integrates well with existing LMMs, consistently improving performance.
\end{abstract}


\renewcommand{\floatpagefraction}{0.8}
\section{Introduction}
\label{sec:intro}
Developments in robotics and embodied AI hold great promise for assisting humans in daily life.
Recent advancements in Large Multimodal Models (LMMs) have significantly accelerated this process, empowering robots with remarkable intelligence in various domains~\cite{khandelwal2022simple, liang2023code, lin2023text2motion, driess2023palm, huang2023visual, yu2023language, yuan2024sd, yuan2025sed}.
Most recently, research has begun to leverage Large Multimodal Models (LMMs) to enable robots to autonomously acquire new skills and experiences in unseen environments, a concept known as Autonomous Skill Acquisition~\cite{zhou2024autonomous, ahn2024autort, yang2024bbsea, katara2024gen2sim, bharadhwaj2024gen2act}.

\begin{figure}[t]
    \centering
    \includegraphics[width=1.0\linewidth]{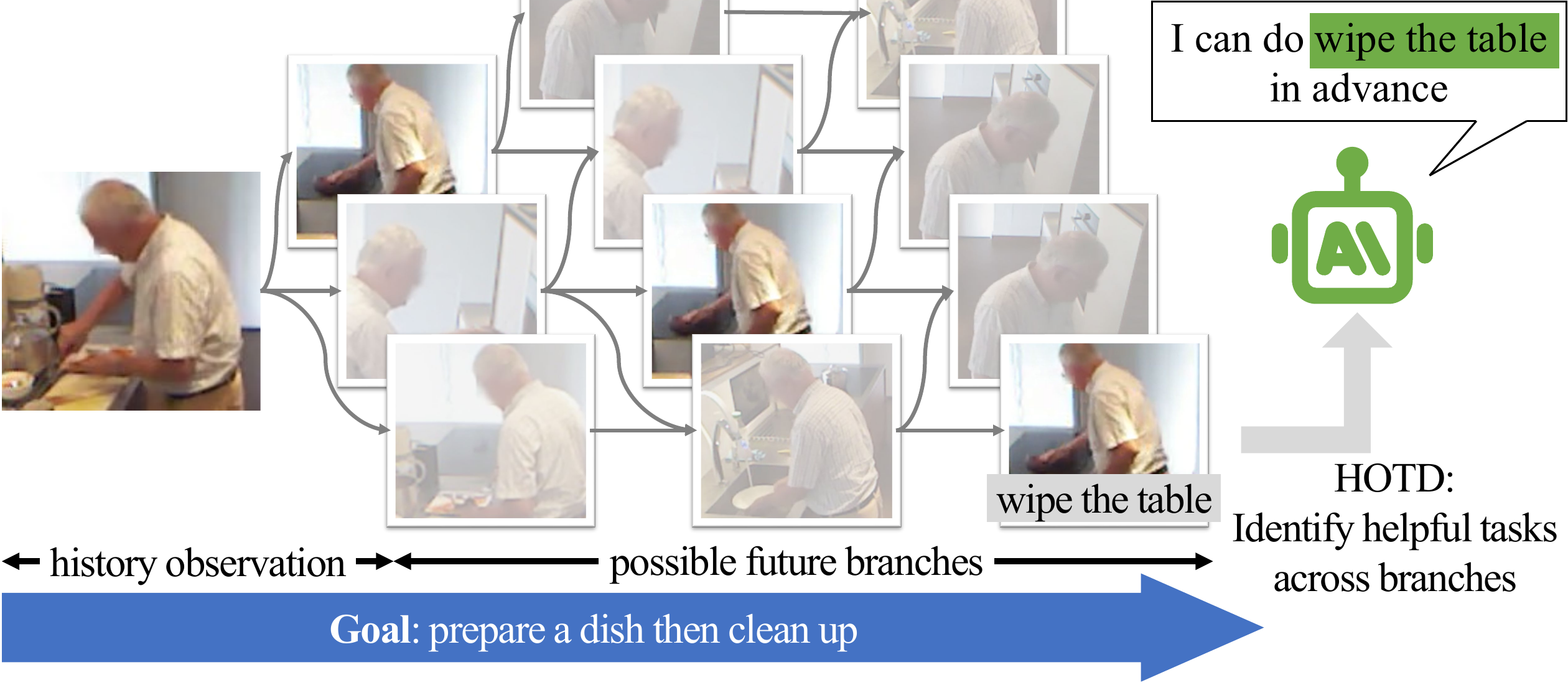}
    \caption{\textbf{The illustration of HOTD.} 
    Driven by an overall goal, humans often engage in concurrent sub-processes, resulting in multiple possible future branches. HOTD aims to identify tasks that remain helpful across diverse and uncertain futures. For example, as highlighted by the green box, completing \emph{wipe the table} in advance saves human effort regardless of the order of other steps.}
    \label{fig:helpful_robot}
\end{figure}

A central capability for autonomous agents is \emph{task discovery}, where LMMs propose manipulation tasks for robots to execute~\cite{wang2023robogen, ahn2024autort, yang2024bbsea, bharadhwaj2024gen2act}. While recent methods focus on generating tasks based on current observations, they typically assume fixed goals or closed environments. However, real-world human contexts are far more complex: people often engage in multiple sub-processes simultaneously, shift intentions dynamically, and rarely make all future steps explicit.
This gives rise to the critical problem of \emph{Human-Centric Open-Future Task Discovery}—inferring tasks that remain helpful across diverse, uncertain future trajectories. Unlike traditional task discovery, which aims to find the next best step toward a known outcome, open-future discovery must anticipate a range of plausible futures and identify actions that support them all. For example, as illustrated in~\cref{fig:helpful_robot}, a robot assisting with housework should proactively wipe the table—a task that remains useful whether the human later cooks, cleans, or rests.
Solving this problem is essential for enabling robots to provide anticipatory, generalizable support in dynamic, human-centered environments. It marks a necessary step toward collaborative AI that is not just responsive, but truly aligned with human intent.


Given these important but neglected aspects, we introduce the novel problem of \textbf{H}uman-centric \textbf{O}pen-future \textbf{T}ask \textbf{D}iscovery.
To clearly study it, we begin by formally defining the HOTD problem, motivated by its central premise: discovering human-centric tasks that arise from open-ended future possibilities.
A formal definition of human-centric tasks is also provided to support this formulation.

However, assessing task-discovery performance under the open-future setting poses significant challenges.
As discussed above, the concurrent nature of human behavior leads to exponential growth in possible future branches, making it prohibitively expensive to annotate the complete set of helpful tasks.
Moreover, relying on human annotators may introduce subjective bias.
To address these issues, we present the HOTD-Bench along with a simulation-based evaluation approach. The HOTD-Bench is sourced from existing datasets and encompasses diverse real-world activities.
The simulation evaluation adopts a discriminative strategy, avoiding exhaustive enumeration while ensuring alignment with human preferences.
By leveraging the world-knowledge in LLMs~\cite{jin2024time, gruver2023large, caotempo}, our evaluation method accommodates open-set hypothetical future branches, including those not realized in the observed scenario.
This mitigates the limitations of purely observational evaluation, enabling comprehensive assessment of both practical and theoretically optimal task decompositions.
Experimental results on the HOTD-Bench reveal that existing LMMs achieve only limited performance on HOTD.

To bridge this gap, we introduce a Collaborative Multi-Agent Search Tree framework (CMAST).
Our core idea is to construct the search tree of procedural structure and identify appropriate tasks accordingly.
Our framework has two major innovations.
First, CMAST introduces a search tree module to explicitly structure the reasoning process.
By iteratively building the search tree, our model captures the inherent uncertainty of action procedures, thoroughly exploring various future scenarios.
Furthermore, the search tree module allows for a scalable test-time thinking, a key trait shared with OpenAI-O3~\cite{openai2025o3minisystemcard} and DeepSeek-R1~\cite{guo2025deepseek}.
Second, CMAST employs a collaborative multi-agent system, where specialized agents manage different stages of the reasoning.
This collaboration effectively decomposes the complex reasoning, enabling each agent to focus on a specific aspect, reducing difficulty.

Experimental results show that our framework significantly outperforms existing LMMs in terms of Valid Task Ratio, while maintaining competitive performance in Valid Task Count, demonstrating its strong advantage in the HOTD.
Ablation studies confirm the effectiveness of the search tree module and demonstrate that our framework can seamlessly integrate with various LMMs.
Visualizations further illustrate the framework’s ability to suggest appropriate tasks by exploring diverse future procedures.
Additionally, experiments validate the effectiveness of the simulator in reasonably deducing future scenarios.

Our key contributions are as follows:
(1) We introduce and formulate the Human-centric Open-future Task Discovery problem, contributing to effective human-AI collaboration.
(2) We establish HOTD-Bench, consisting of over 2K real-world videos from two sources.
A simulator is proposed to deduce a given task's future outcome, enabling the assessment of any future trajectory and the evaluation of its helpfulness.
(3) We propose the CMAST framework to manage the complex reasoning.
Experiments show that CMAST framework can seamlessly integrate a variety of existing LMMs and achieve consistently superior performance.

\section{Related Work}
\label{sec:related_work}

\begin{figure*}[t]
    \centering
    \includegraphics[width=0.8\linewidth]{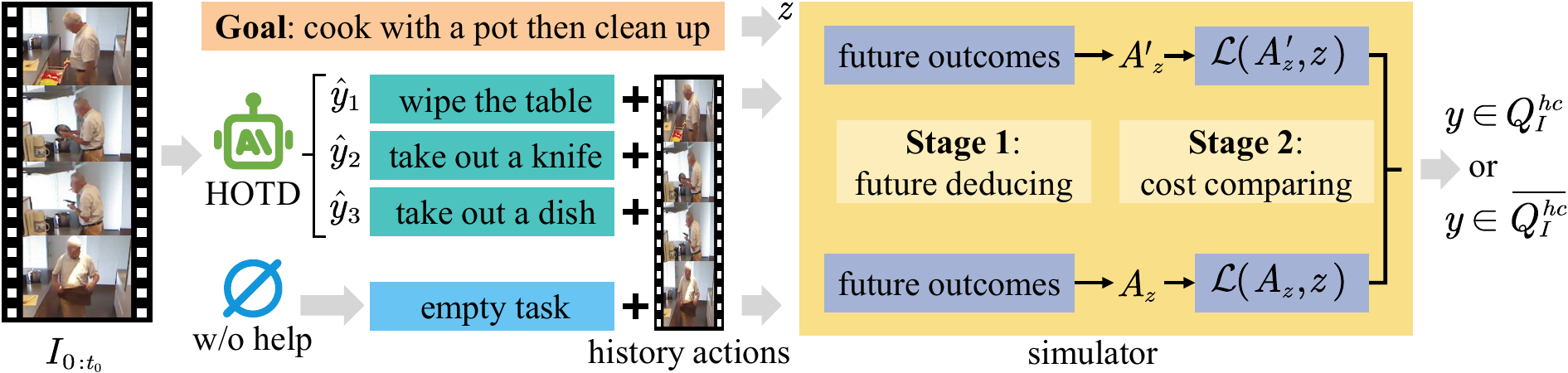}
    
    \centering
    \includegraphics[width=0.92\linewidth]{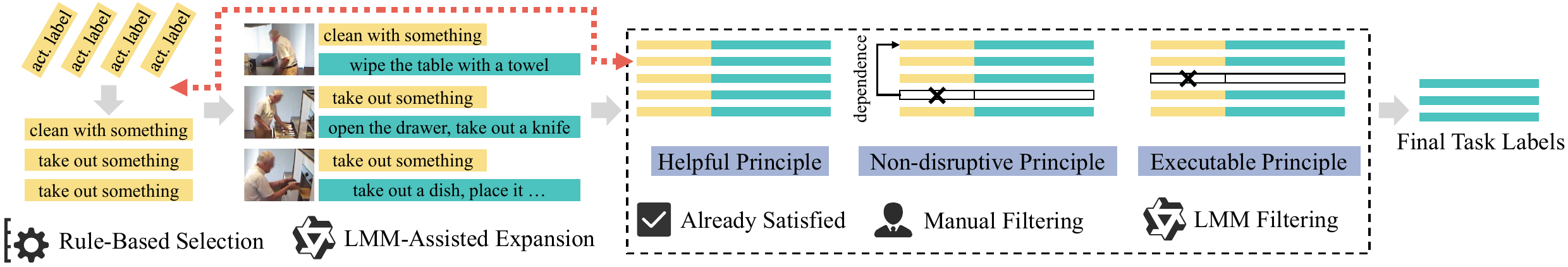}

    \caption{\textbf{The process of simulation-based evaluation (upper) \& annotation pipeline (lower).} The simulator first takes a discovered task $\hat{y}_n$, a historical action sequence and its associated goal $z$ as input. Then, it simulates the resulting future trajectory $A^{\prime}_z$ by accounting for the adjusted human actions until the goal. Finally, it summarizes the overall process to estimate the incurred cost $\mathcal{L}(A^{\prime}_z, z)$ and compares it to the original cost $\mathcal{L}(A_z, z)$.
    In the annotation pipeline, future actions are first selected to meet the \emph{helpful principle}, then expanded into descriptive sentences and filtered through the \emph{non-disruptive principle} and the \emph{executable principle}, finally forming the task labels.}
    \label{fig:simu_process}
\end{figure*}

\noindent \textbf{Autonomous Skill Acquisition.}
This technology encourages robots to learn new skills with less human instruction~\cite{bharadhwaj2024gen2act, ahn2024autort}.
Some studies focus on simulation-based learning~\cite{wang2023robogen, wang2023gensim, zhao2024agentic, katara2024gen2sim, yang2024bbsea}, leveraging LMMs to automatically generate simulation components, such as scene assets and supervision.
Other studies focus on real-world learning~\cite{ahn2024autort, zhou2024autonomous}.
They typically leverage LMMs to recommend physical-interactive tasks, allowing the robot to attempt them and gather experience.
In this work, we aim to further enhance LMM’s task-proposing capabilities, focusing on generating more valuable tasks to assist humans.
Through this enhancement, we hope robots will acquire skills that directly align with human needs.

\noindent \textbf{Enhancing LMMs for Complex Reasoning.}
Recent studies have focused on enhancing LMMs for complex reasoning, with two common approaches being Multi-Agent Systems and Chain-of-Thought.
Multi-agent system deploys multiple agents to break down complex problems into smaller, manageable sub-parts~\cite{wu2023autogen, hong2023metagpt, yuan2025video, yu2025fincon}, which has been verified across various applications~\cite{yang2024rila, yang2024embodied, aher2023using}.
Chain-of-thought reasoning enhances LMMs by generating intermediate steps that progressively lead to the final answer~\cite{guo2025deepseek, yuan2025autodrive}.
The most notable models are DeepSeek-R1~\cite{guo2025deepseek}, OpenAI-o1/o3~\cite{openaio1, openai2025o3minisystemcard}, which feature a \emph{scalable test-time thinking} that performs increasingly efficacious CoT reasoning with extended thinking time.
Inspired by previous works, our approach combines the advantages of both approaches.

\noindent \textbf{Related Video Datasets.}
Video understanding has always been a research focus.
There exist a number of video understanding datasets~\cite{soomro2012ucf101, caba2015activitynet, jia2020lemma, damen2022rescaling, zeng2024benchmarking, zhou2023procedure}, some of which are related to humans\cite{zellers2019recognition, lei2018tvqa, grauman2022ego4d}.
However, none of the existing datasets focus on task discovery, where the answers are supportive tasks for humans.
In this work, we curate our benchmark based on the existing video datasets and develop an evaluation approach to assess the contribution of any discovered task.
\section{Formulation}


\subsection{Problem Formulation}

The HOTD problem aims to discover a set of tasks that assist the human in a given video segment. Specifically, given an input video segment $I_{0:t_0}$, the model generates a set of predicted tasks, denoted as $\hat{Q}_I=\left\{ \hat{y}_1,\hat{y}_2, \cdots,\hat{y}_i \right\}$, where $\hat{\phantom{x}}$ indicates the predicted value.

To formally define the objective of HOTD, we denotes $Q$ be the set of all task descriptions in natural language.
For a given video input $I$, we define $Q^{hc}_{I}$ as the subset of $Q$ that contains only the human-centric tasks constrained by the observed scenario $I$. An HOTD method is characterized by a mapping function $G$, which predicts a set of tasks based on the given video input: $\hat{Q}_I = G(I).$
The objective of HOTD is to find an optimal mapping function $G$ that maximizes the inclusion of human-centric tasks:
\begin{equation}
    \underset{G}{\max } \, |\hat{Q}_I \cap Q^{hc}_I|, \,\,\,\,
    \underset{G}{\max } \, \frac{|\hat{Q}_I \cap Q^{hc}_I|}{|\hat{Q}_I|},
    \label{eq:obj}
\end{equation}
where the second is the normalized version of the first. $\hat{Q}_I$ denotes the predicted task set, $Q^{hc}_I$ denotes the ground truth human-centric task set. We further use $\overline{Q^{hc}_I}$ to denote the complement of $Q^{hc}_I$, only containing unhelpful tasks.

For the notion of open-ended future possibilities, we do not explicitly define it.
Instead, we rely on the simulation evaluation, which accommodates open-set future scenarios and estimates outcomes in line with human preferences.


\subsection{The Definition of Human-Centric Tasks}
\label{sec:task_define}
To establish a clear understanding of HOTD, we develop a systematic definition of what human-centric tasks are.
Concretely, a human-centric task is defined as an executable action that contributes to achieving the human's goal.

First, each human-centric task corresponds to a specific action that can be executed by robots. We follow the three-level taxonomy, `action primitive', `action', and `activity', introduced by Moeslund et al.~\cite{moeslund2006survey}.
The human-centric task are defined to be at the second level, ensuring that they are neither too narrow to lack meaning nor too broad to confuse the robot.

Second, a human-centric task is not merely an action but one that assists a human in reaching their goal. 
Let $z \in Z$ denote a human's latent goal. The human executes a sequence of actions to accomplish $z$, denoted as $A_z = \left\{a_1, a_2, \cdots, a_n\right\}$. When the robot performs an additional task $y$, the human may adjust their actions accordingly, resulting in a modified action sequence $A^{\prime}_z = \left\{y, a^{\prime}_1, a^{\prime}_2, \cdots, a^{\prime}_n\right\}$, where $a^{\prime}_n$ represents the human's modified action due to the influence of the robot’s intervention.

To formally define whether $y$ provides assistance, we introduce a cost function $\mathcal{L}$ that quantifies the cost required to achieve $z$. This cost function may be defined in terms of time spent, labor exerted, or other measures. A task $y$ is considered helpful if its inclusion reduces the overall cost, where the subscript $I$ denotes the history video driven by $z$:
\begin{equation}
    y \in Q^{hc}_{I} \Longleftrightarrow y \in Q \land \mathcal{L}(A^{\prime}_z, z) < \mathcal{L}(A_z, z).
    \label{eq:cost_func_a}
\end{equation}
Conversely, a task that increases the cost is not helpful:
\begin{equation}
    y \in \overline{Q^{hc}_{I}} \Longleftrightarrow y \in Q \land \mathcal{L}(A^{\prime}_z, z) \geqslant \mathcal{L}(A_z, z).
    \label{eq:cost_func_b}
\end{equation}
Such a discriminative definition is particularly appropriate and necessary, as the complexity of human cognition makes a prescriptive one infeasible.

\section{Benchmark}
\label{sec:bench}
\subsection{Data Collection}
In order to evaluate the model's performance under practical scenarios, we construct our dataset from two existing datasets: the Toyota Smarthome Untrimmed (TSU)~\cite{das2019toyota, dai2022toyota} and the Charades (CHA)~\cite{sigurdsson2016hollywood}.
They offer various real-world activities from two distributions.
We apply a sliding window to segment the videos and filter out low-quality samples, resulting in 2450 curated clips totaling nearly 40 hours. The TSU and CHA subsets contribute 2K and 0.4K videos, respectively.

\subsection{Evaluate by Simulation}
Under the open future setting, our evaluation aims to quantify how many human-centric tasks are included in the predicted set. 
However, such evaluation is nontrivial.
A straightforward approach would be to have human annotators label all helpful tasks.
While intuitive, it is impractical due to the annotator subjectivity and the prohibitive cost of exhaustively labeling exponentially many future branches (see~\cref{fig:helpful_robot}).
Instead, verifying whether a given task is helpful is substantially more tractable, as it only requires estimating the cost introduced in~\cref{eq:cost_func_a} and~\cref{eq:cost_func_b}.
This motivates our use of simulation as an evaluation tool.
A simulator can flexibly model how the future would unfold under any hypothetical task insertion, thereby enabling the evaluation of arbitrary candidate trajectories.

The proposed simulation-based evaluation approach is illustrated in the upper part of~\cref{fig:simu_process}.
For sequences without robot intervention, it directly simulates the future processes. For sequences with robot intervention, it models human adaptation and reconstructs the complete sequence.
The resulting trajectory is then used to estimate the incurred costs.
A central strength of this approach lies in its generative nature, which allows for evaluating not only observed trajectories but also any hypothetical future beyond the dataset, thereby capturing scenarios that could be more optimal than those explicitly performed.

In our implementation, we adopt an LLM as the simulator, which has been proven to reliably deduce future evolution~\cite{jin2024time, gruver2023large, caotempo}, while also aligning with human preferences and minimizing subjective bias~\cite{rafailov2023direct, bai2022constitutional}.
The latent goal $z$ is pre-annotated according to the whole video.
The cost is defined in terms of time consumption.
To mitigate the sensitivity caused by absolute time estimations, our simulator evaluates the relative time costs by comparing two action sequences with or without robot intervention.
We adopt a chain-of-thought prompting approach that first guides the LLM to envision future outcomes and then determines whether $\mathcal{L}(A^{\prime}_z, z) < \mathcal{L}(A_z, z)$.
The experiment in Sec. 6.3 as well as~\cref{fig:alignemnt} and~\cref{fig:simulator_sample} validates that our LLM simulator is simple yet reliable.

\subsection{Evaluate by Labels}
\label{sec:eval_by_labels}
We introduce an alternative evaluation approach based on labeled task sets $\tilde{Q}^{hc}_{I} \approx Q^{hc}_{I}$, enabling a more stable evaluation.
The core idea is to approximate a ``ground truth'' set of human-centric tasks, assuming that tasks humans will inevitably perform are inherently beneficial if completed in advance, as they directly reduce human effort and cost.

To construct $\tilde{Q}^{hc}_{I}$, we design a semi-automated annotation pipeline (see~\cref{fig:simu_process} lower).
We begin with the dense action labels from the original datasets, select the future-performed actions, and expand them into descriptive, open-set task sentences using Qwen-VL~\cite{wang2024qwen2}. The annotation is guided by three predefined principles defining human-centric tasks.
\emph{Helpful}: A human-centric task should complete actions in advance that humans would otherwise have to do themselves. 
\emph{Non-disruptive}: A human-centric task should not conflict with the human's plan.
\emph{Executable}: A human-centric task must be possible to execute given the current conditions.
The detailed implementation of them can be found in Appendix.
By filtering and refining the expanded action descriptions under these principles, we obtain a reliable approximation of the ground truth task set.


\begin{figure}[t]
    \centering
    \includegraphics[width=1.0\linewidth]{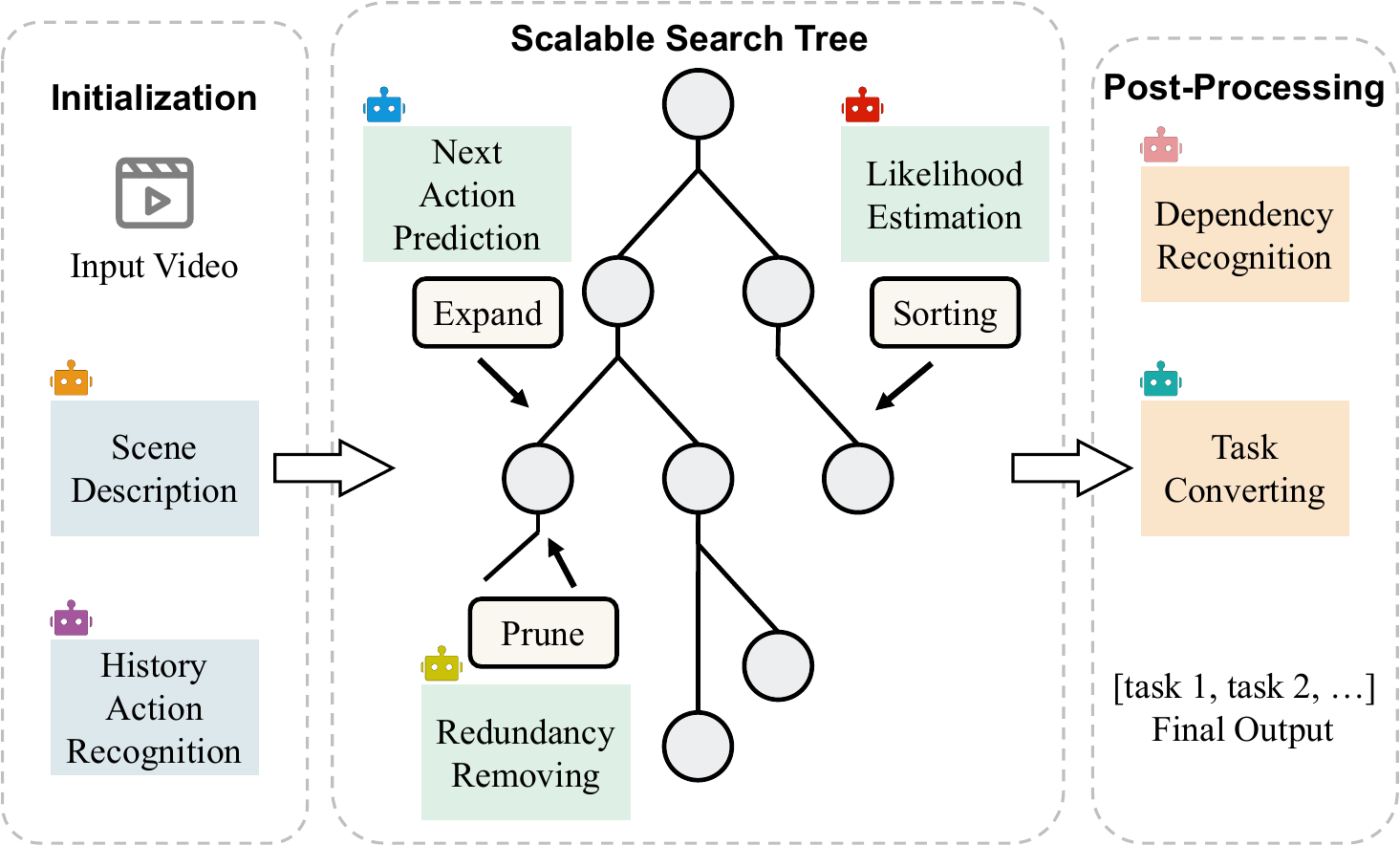}
    \caption{\textbf{The overview of the Collaborative Multi-Agent Search Tree framework.} It structures the HOTD reasoning with 7 LMM agents and a scalable search tree module.}
    \label{fig:search_tree_model}
\end{figure}

\section{Method}
\label{sec:cmast_method}

\subsection{Search Tree Module}

The HOTD reasoning not only requires understanding the visual contents, but also analyzing the open-future procedure.
To achieve this, we build a search tree that explicitly represents open-future action space, enabling an explicit exploration of future scenarios.
Additionally, it supports flexible expansion and pruning, facilitating \emph{scalable test-time thinking} that achieves comprehensive answers, a key trait shared with OpenAI-O3~\cite{openai2025o3minisystemcard} and DeepSeek-R1~\cite{guo2025deepseek}.

The search tree is comprised of a set of nodes and edges:
$T=\left( V,E \right)$.
Each node corresponds to an individual action.
The edge between nodes represents the temporal order.

The first N nodes in the tree represent the history action sequence determined by the input video, formulated as:
\begin{equation}
\begin{gathered}
    T^{0:N}=P\left(v^{0}, v^{N} \right) =\left( v^{0},v^{1}, \cdots, v^{N} \right) ; \\ \forall n<N,\left| Child\left( v^n \right) \right|=1 .
    \label{eq:first_n}
\end{gathered}
\end{equation}
The superscript $n$ represents the n-th layer of the tree.
The symbol $P\left(v, v\prime \right)$ represents the path from node $v$ to node $v\prime$.
The $Child\left( v \right)$ represents the child nodes of the node $v$.

Starting from the N-th node, the tree begins to branch out.
Each branch represents a possible next step, reflecting the uncertainty of the open-future, formulated as:
\begin{equation}
    Child\left( v^n \right) =g\left( v^n \right) =\left\{ v_1,v_2,\cdots,v_k \right\} , n > N.
    \label{eq:branch}
\end{equation}
where $g$ is a node expansion function which we will introduce in~\cref{eq:expand}.
As the tree progresses, multiple branches capture various possible action sequences.
The leaf nodes in the tree represent the completion of the entire activity, which we refer to as the `finish activity'.

The search tree supports several searching strategies. To balance performance and efficiency, we employ a pruned exhaustive search with a 0.5 probability threshold. Alternative strategies are discussed in the Sec. 6.3.

\subsection{Collaborative Multi-Agent System}
\label{subsec:mac}


Inspired by previous works~\cite{wu2023autogen, hong2023metagpt, yu2025fincon}, we develop a multi-agent system to structure the reasoning process in the HOTD problem.
The workflow is presented in \cref{fig:search_tree_model}.
Our key idea is to decompose the complex problem by aligning each agent with a specific stage of a search tree, such as initialization, expansion, pruning, and post-processing. This design not only preserves the generalization ability of LMMs but also enables seamless integration with various LMMs.


First, receiving the input video, the \emph{Scene Description Agent} digests the video and produces a detailed description $s = f_s(I_{0:t_0})$, providing an overall precondition.

Second, the \emph{History Action Recognition Agent} recognizes the history actions of the person, $\left( v^0,v^1,\cdots,v^N \right) =f_r\left( I_{0:t_0}, s \right)$, which initializes the search tree.

Third, from the initial search tree, three agents are employed to expand the search tree iteratively.
In each iteration, the \emph{Next Action Prediction Agent} forecasts the next immediate action given the entire action path up to now.
And the \emph{Likelihood Estimation Agent} predicts the probability for each child node, providing reference for sorting and pruning.
We manually add the `finish activity' node to every non-leaf node.
When the above two agents expand the search tree, the \emph{Redundancy Removing Agent} is employed to prune the redundant nodes.
These three agents work iteratively until all unexpanded nodes are leaf nodes or reach a maximum tree height.
Let $g$ denote the above three agents together, and the expansion process can be written as:
\begin{equation}
    \left\{ v_1,v_2,\cdots,v_k \right\} =g\left( v^n, I_{0:t_0}, s \right) ,n>N.
    \label{eq:expand}
\end{equation}



Fourth, the search tree is formatted into a set of action sequences by traversing all the paths.
The \emph{Dependency Recognition Agent} is adopted to identify and exclude actions with prerequisite, retaining only those executable ones.

Finally we dismantle the sequential structure, yielding a set of independent actions. The \emph{Task Converting Agent} is employed to transform each action into a task description from the robot’s perspective, formulated as:
\begin{equation}
\begin{aligned}
    \hat{Q}_I&=\left\{ \hat{y}_1,\hat{y}_2,\cdots,\hat{y}_i \right\} \\&= f_c\left( \left\{ v|v\in \cup_{n>N}{V^n} \right\}, I_{0:t_0}, s \right) ,
    \label{eq:convert}
\end{aligned}
\end{equation}

\subsection{Implementation Details}
Our framework is entirely training-free, requiring no fine-tuning of the overall system or any submodule. This design allows seamless integration of various LMMs. Specifically, the \emph{Scene Description Agent}, the \emph{History Action Recognition Agent}, and the \emph{Next Action Prediction Agent} are LMM agents, implemented with LLaVA-Next-Video~\cite{zhang2024llavanextvideo}, the other three agents are are LLM agents, implemented with Qwen-LM~\cite{yang2024qwen2technicalreport}.

\begin{table*}
    \centering
    \begin{adjustbox}{scale=0.78}
    \setlength{\tabcolsep}{4pt}
    \begin{tabular}{@{}l|cccc|cc@{}}
      \toprule
       & \multicolumn{4}{c|}{TSU subset} & \multicolumn{2}{c}{CHA subset} \\
      Method & vc@20 / vr@20 & vc@40 / vr@40 & vc@60 / vr@60 & vc@80 / vr@80 & vc@20 / vr@20 & vc@30 / vr@30 \\
      \midrule
Qwen2VL-7B    & 2.86 / 44.3\% & 2.71 / 44.2\% & \textbf{4.03} / 40.6\% & 2.77 / 42.0\% & 2.06 / 43.1\% & 2.08 / 48.1\% \\
Qwen2.5VL-72B     & 2.45 / 48.8\% & 2.47 / 47.6\% & 2.62 / 42.6\% & 2.22 / 43.0\% & 3.01 / 40.9\% & 2.87 / 43.0\% \\
InternVL2-8B    & 2.23 / \underline{58.4}\% & 2.51 / \underline{61.0}\% & 2.44 / \underline{62.4}\% & 2.79 / \underline{59.8}\% & 2.47 / \underline{54.5}\% & 2.28 / \textbf{58.1}\% \\
InternVL2.5-26B   & 1.57 / 40.5\% & 1.48 / 42.4\% & 1.42 / 40.9\% & 1.38 / 40.5\% & 1.88 / 42.2\% & 1.74 / 43.3\% \\
Video-LLaVA-7B  & 0.47 / 40.8\% & 0.45 / 41.5\% & 0.40 / 40.6\% & 0.46 / 42.1\% & 0.36 / 36.0\% & 0.29 / 29.1\% \\
LLaVA-NV-7B  & \underline{3.62} / 48.9\% & 3.34 / 50.2\% & 2.92 / 48.8\% & \underline{3.41} / 51.5\% & \textbf{6.20} / 54.1\% & \textbf{6.02} / 50.0\% \\
LLaVA-NV-34B & 3.28 / 41.5\% & \underline{3.39} / 44.2\% & 2.69 / 45.0\% & \underline{3.41} / 42.6\% & \underline{3.55} / 40.8\% & \underline{3.18} / 42.7\% \\
    \bottomrule
CMAST(ours)  & \textbf{3.90} / \textbf{71.7}\% & \textbf{3.83} / \textbf{71.9}\% & \underline{3.89} / \textbf{72.0}\% & \textbf{3.86} / \textbf{71.8}\% & 2.73 / \textbf{55.5}\% & 2.82 / \underline{53.6}\% \\
      \bottomrule
    \end{tabular}
    \end{adjustbox}
    \caption{\textbf{The quantitative comparison in the HOTD-Bench, evaluated by simulation.} The `vc@' and `vr@' denotes the Valid Task Count and the Valid Task Ratio at a specific observation length, higher is better. The best are highlighted by \textbf{bold} and the second best are highlighted by \underline{underline}. CMAST demonstrates significantly better performance against other methods.}
    \label{tab:comparison}
\end{table*}

\begin{table}
    \centering
    \begin{adjustbox}{scale=0.78}
    \setlength{\tabcolsep}{4pt}
    \begin{tabular}{@{}l|cc|c@{}}
      \toprule
       & \multicolumn{2}{c|}{TSU subset} & \multicolumn{1}{c}{CHA subset} \\
      Method & vc@40 / vr@40 & vc@60 / vr@60 & vc@20 / vr@20 \\
      \midrule
Qwen2VL-7B     & 0.82 / 14.1\% & 0.82 / 13.9\% & 0.58 / \underline{13.2}\% \\
InternVL2-8B    & 0.50 / 16.0\% & 0.52 / 17.1\% & \underline{0.89} / ~~5.3\% \\
Video-LLaVA  & 0.22 / \underline{20.6}\% & 0.22 / \underline{21.5}\% & 0.09 / ~~9.2\% \\
LLaVA-NV-7B  & \underline{1.51} / 15.0\% & \underline{1.61} / 15.0\% & \textbf{1.00} / ~~8.1\% \\
LLaVA-NV-34B & 1.22 / 13.1\% & 1.30 / 13.8\% & 0.87 / 10.9\% \\
    \bottomrule
CMAST(ours)  & \textbf{1.92} / \textbf{38.7}\% & \textbf{1.98} / \textbf{39.3}\% & 0.79 / \textbf{15.6}\% \\
      \bottomrule
    \end{tabular}
    \end{adjustbox}
    \caption{\textbf{The quantitative comparison in the HOTD-Bench}, evaluated by labels. CMAST also demonstrates significantly better performance against other methods.}
    \label{tab:comparison_label}
\end{table}

\begin{figure*}[t]
  \centering
  \begin{minipage}[b]{0.37\linewidth}
    \centering
    \includegraphics[width=0.9\linewidth]{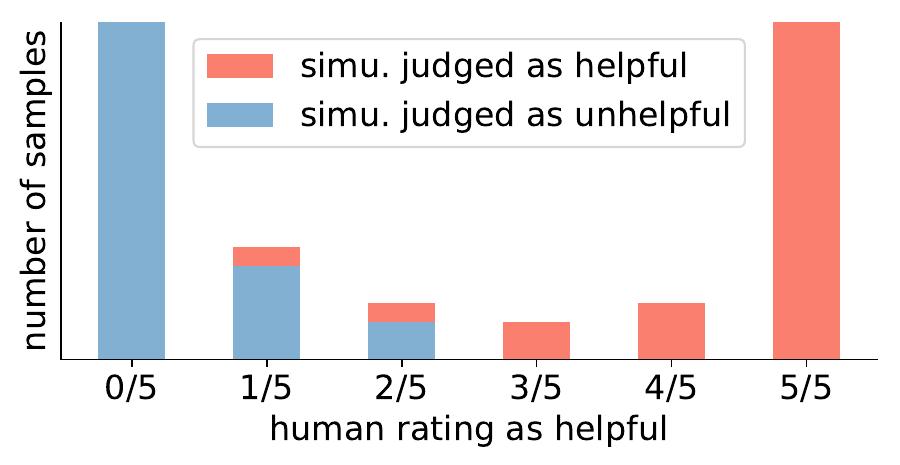}
    \vspace{0.1cm}
    \caption{\textbf{Human evaluation of the simulator.} Columns indicate how many annotators rated each task as helpful, where 0 means all rated it unhelpful and 5 means all rated it helpful. The distribution shows strong agreement between the simulator and human preferences.}
    \label{fig:alignemnt}
  \end{minipage}
  \hfill
  \begin{minipage}[b]{0.30\linewidth}
    \centering
    \includegraphics[width=\linewidth]{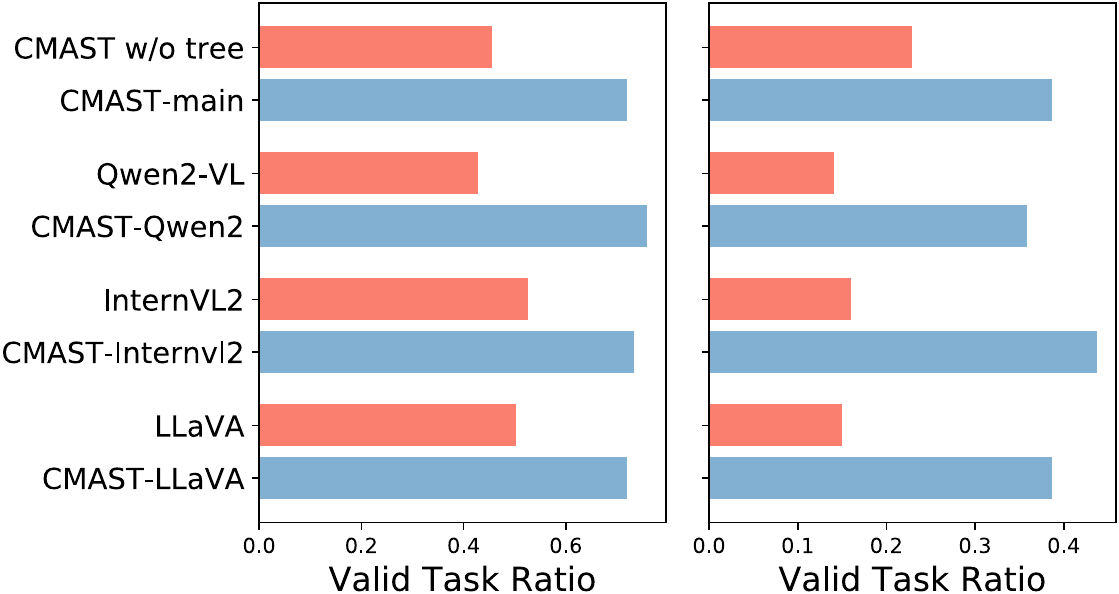}
    \vspace{0.0cm}
    \caption{\textbf{The ablation studies} on the search tree module and the component agents. We show the results given 40 sec of observations in the TSU, evaluated by simulator (left) and labels (right).}
    \label{fig:ablation_agent}
  \end{minipage}
  \hfill
  \begin{minipage}[b]{0.30\linewidth}
    \centering
    \includegraphics[width=\linewidth]{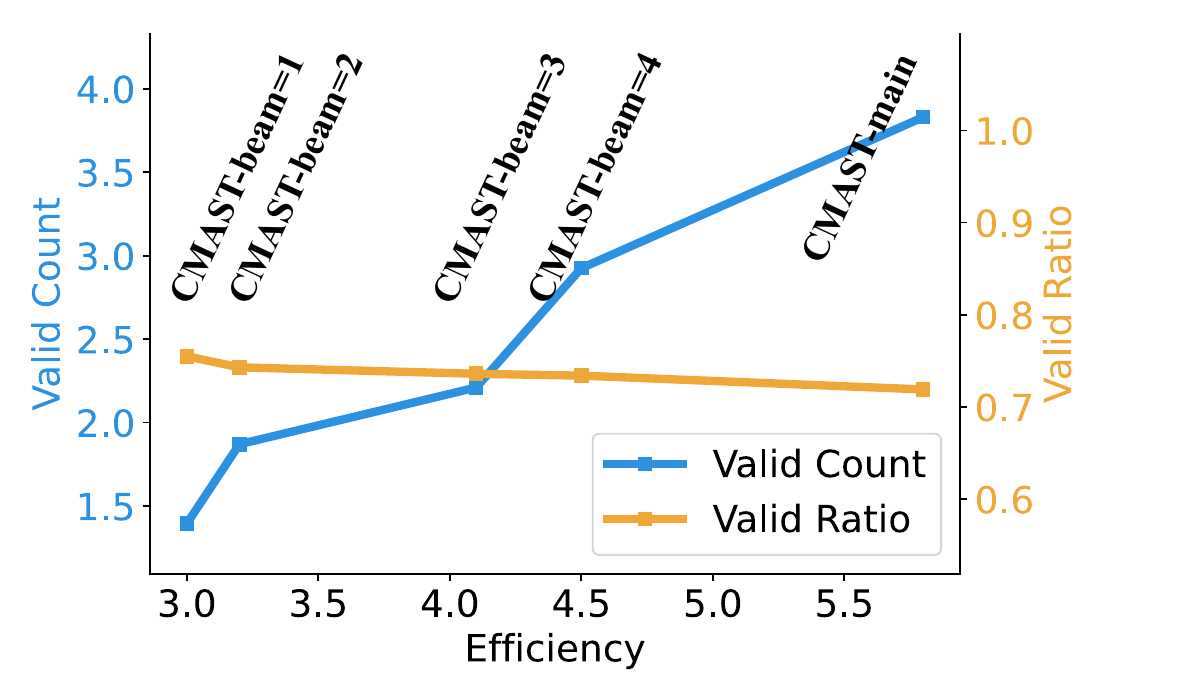}
    \vspace{0.0cm}
    \caption{\textbf{The ablation study} on the search strategy in the TSU subset, 40 sec observation length, evaluated by simulation. The `Efficiency' is measured by the average number of expansions.}
    \label{fig:ablation_search}
  \end{minipage}
\end{figure*}

\section{Experiment}
\label{sec:experiment}

\subsection{Experimental Setups}
\noindent \textbf{Baselines.}
We conduct a quantitative study on the HOTD-Bench to evaluate the HOTD capabilities of the following open-source LMMs:
Qwen2-VL~\cite{wang2024qwen2}, Qwen2.5-VL~\cite{qwen2.5-VL}, InternVL2~\cite{internvl2}, InternVL2.5~\cite{chen2024expanding}, Video-LLaVA~\cite{lin2023video}, LLaVA-Next-Video~\cite{zhang2024llavanextvideo}.
For these baselines, we give them the video and directly prompt them to recommend some assisting tasks.

\noindent \textbf{Evaluation Metrics.}
We establish two evaluation metrics, Valid Task Count and Valid Task Ratio, corresponding to two objectives in~\cref{eq:obj}.
The Valid Task Count, denoted as `vc', measures the average number of helpful tasks discovered, indicating the model's ability to identify a diverse range of helpful tasks.
The Valid Task Ratio, denoted as `vr', measures the average proportion of helpful tasks within each prediction, indicating the reliability and the precision of the model's outputs.
Their formulations are shown below, where $N$ denotes the number of samples:
\begin{equation}
    \text{vc} = \frac{1}{N}\sum_{n=1}^N{\left| Q^{hc}_{I_n} \cap \hat{Q}_{I_n} \right|}, \,\, \text{vr} = \frac{1}{N}\sum_{n=1}^N{\frac{\left| Q^{hc}_{I_n} \cap \hat{Q}_{I_n} \right|}{\left| \hat{Q}_{I_n} \right|}}.
    \label{eq:metrics}
\end{equation}

For the majority of experiments in this paper, we report the results of both evaluation approaches (by simulation and labels). We also measured videos with different observation lengths, using `@' for distinguishing. For example, `vc@40' represents the metric given 40 sec of observations.

\subsection{Overall Evaluation}

\cref{tab:comparison} shows the performance of our framework against the baseline LMMs. Our method demonstrates significant advantages over baseline methods, validating its effectiveness.

For the Valid Task Ratio, our method significantly outperforms the other methods for all the observation lengths.
Specifically, all existing LMMs achieve a relatively low Valid Task Ratio, ranging from 29.1\% to 62.4\%, struggling to provide sufficient assistance.
We suspect this is due to their training on dialogue instruction corpora, which is inadequate to capture the behavioral expectations of humans.
In contrast, our approach surpasses the second-best method by 15\% to 22\% in the TSU subset and achieves the same level as the second-best method in the CHA subset.
This validates that by forecasting future situations and imagining helpful tasks in the future context, our method can discover more supportive tasks within each prediction round.

For the Valid Task Count, our method also achieves competitive results.
Our method outperforms the second-best by an average of 7.6\% in the TSU subset.
This confirms that our method is capable of discovering a diversity of tasks, with a significant proportion of them being beneficial, largely due to the search tree module's exploration of potential future scenarios and activity procedures.

\subsection{Further Analysis}
\label{sec:further_analysis}

\noindent \textbf{Comparing Among Existing LMMs.}
Based on the results in \cref{tab:comparison}, we further analyze the performance of existing LMMs. Notably, there is generally a trade-off between Valid Task Count and Valid Task Ratio.
For instance, Internvl2-8B~\cite{internvl2} achieves the highest Valid Task Ratio among baselines, but a low Valid Task Count, indicating overly conservative predictions.
LLaVA-Next-Video-7B~\cite{zhang2024llavanextvideo} achieves the highest Valid Task Count among baselines, but a low Valid Task Ratio, indicating aggressive predictions with many invalid tasks.
Moreover, larger models offer no clear advantage over smaller models across our experiments.
This indicates that scaling up does not necessarily enhance task discovery capabilities in this setting.
Overall, current LMMs still face notable challenges in addressing the HOTD problem.

\noindent \textbf{Reliability of the Simulator.}
To investigate \emph{whether the simulator aligns with human preferences}, we conducted a human evaluation. We randomly selected 25 helpful and 25 unhelpful tasks as marked by the simulator, and presented them in mixed order to 5 annotators. Annotators independently judged whether each task is helpful. As shown in ~\cref{fig:alignemnt}, tasks judged helpful by the simulator were largely considered helpful by humans, and similarly for unhelpful tasks. The result shows that the simulator reliably reflects human preferences, offering accurate assessments.

\begin{figure*}
    \centering
    \includegraphics[width=0.95\linewidth]{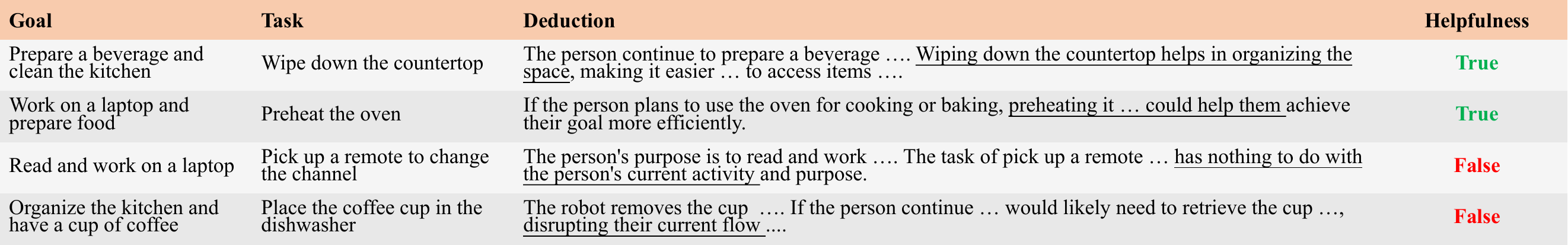}
    \caption{\textbf{Examples of the simulator's future deduction and helpfulness judgment.}}
    \label{fig:simulator_sample}
\end{figure*}

\noindent \textbf{Case Study of the Simulator.}
~\cref{fig:simulator_sample} presents future deductions generated by our simulator, demonstrating its ability to model human actions in response, even for scenarios not explicitly observed in the dataset.
For instance, in the last row, it successfully anticipates that the person will need to retrieve the cup as a consequence of the robot’s action, an event that wouldn’t have happened otherwise.
These examples highlight the effectiveness of our LLM-based simulator, as it offers a reliable and comprehensive assessment of all hypothetical tasks.

\begin{figure}[t]
    \centering
    \includegraphics[width=0.94\linewidth]{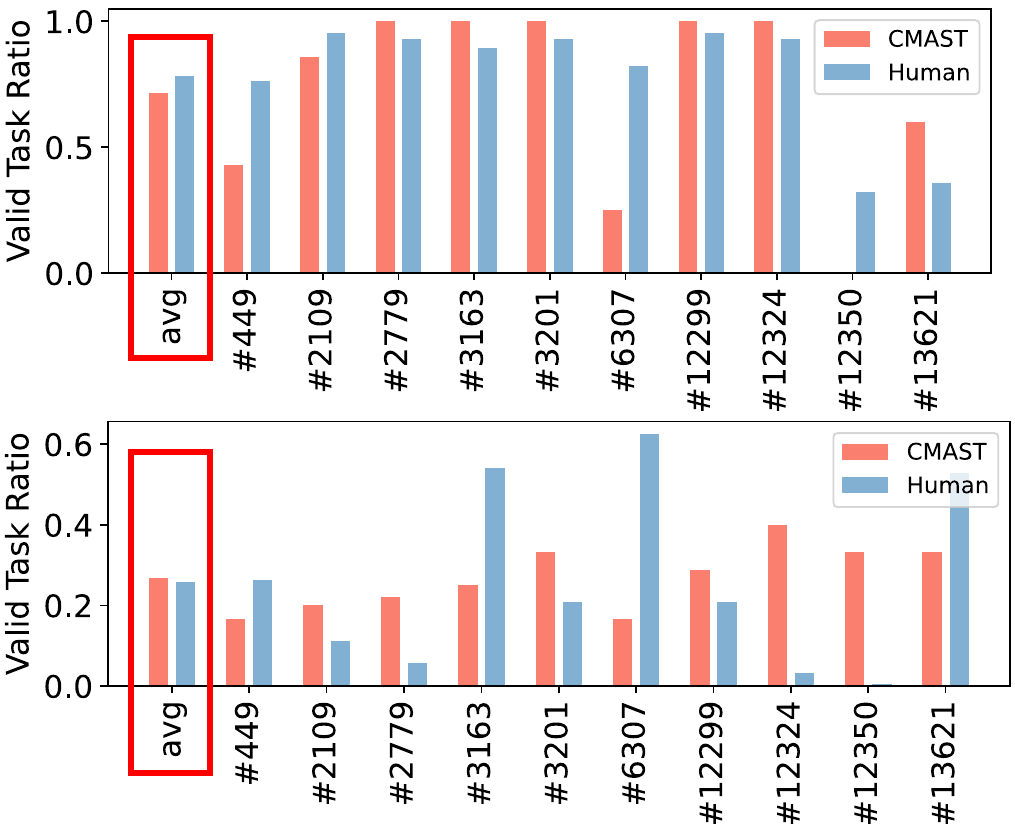}
    \caption{\textbf{Comparisons against human performance.} We show the Valid Task Ratio of 10 samples and the average score. The upper row is evaluated by simulation and the lower row is by labels.}
    \label{fig:human}
\end{figure}

\noindent \textbf{Ablation on the Search Tree Module.}
We conduct an ablation study to examine the overall contribution of the search tree module.
The results are shown in  \cref{fig:ablation_agent}.
The `CMAST-main' is our main version.
The `CMAST w/o tree' removes the entire search tree module, replacing it with an LMM agent that directly predicts future actions.
As can be seen from the results, replacing the entire search tree module reduces the Valid Task Ratio by 37\%.
This suggests that a single LMM agent may produce confined prediction paths.
On the contrary, the search tree module addresses it by providing an explicit and structured procedure space, allowing for a thorough examination of different action procedures.

\noindent \textbf{Ablation on the Search Strategy.}
The results in \cref{fig:ablation_search} compare different search strategies, where `CMAST-beam=$k$' denotes the beam search strategy with a beam size of $k$.
While the greedy search strategy (beam=1) is the most efficient, it discovers only 1.4 helpful tasks per video. As we adopt increasingly time-intensive search strategies, the model demonstrates progressively better performance, with the number of discovered tasks \emph{\textbf{significantly}} increasing while the accuracy ratio remains basically unchanged.
This suggests that by scaling up the test-time thinking, our model is able to explore a broader range of future situations and cover more valuable tasks.



\noindent \textbf{Choice of Component Agents.}
We further conduct experiments to investigate the detailed influence of choosing different component agents.
Specifically, we replace the LMM agents in our framework with different LMMs, generating the following variants:
CMAST-LLaVA, which is our main version,
CMAST-InternVL2, which uses InternVL2-8B~\cite{internvl2},
CMAST-Qwen2, which uses Qwen2-VL-7B~\cite{wang2024qwen2}.
Additionally, we report the corresponding vanilla LMMs for each variant.
The results are shown in \cref{fig:ablation_agent}.
Comparing the same LMMs used in isolation, the LMMs within our framework improve the Valid Task Ratio by at least 39\%.
The results confirm that using different component agents consistently enhances our model's performance, highlighting its ability to seamlessly integrate with various LMMs.

\begin{figure}[t]
    \centering
    \includegraphics[width=1.0\linewidth]{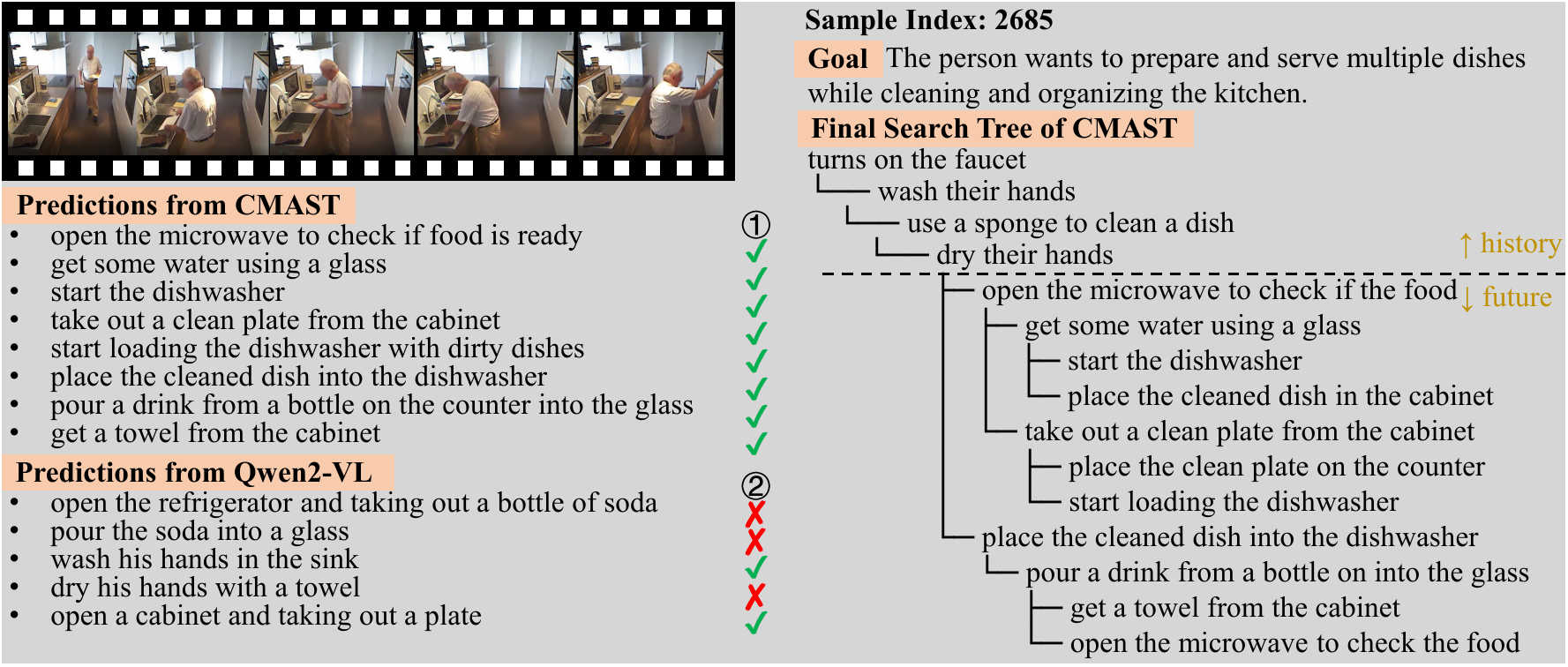}
    \caption{\textbf{The visualization} of predictions from CMAST and baseline~\cite{wang2024qwen2}. The \ding{51} / \ding{55} column indicates correct/incorrect predictions. More examples can be found in Appendix.}
    \label{fig:pred_sampe_main}
\end{figure}

\noindent \textbf{Comparing with Human Performance.}
This part investigates \emph{whether CMAST can achieve human-level performance}.
To this end, we randomly select 10 examples and ask human participants to discover tasks.
The results are in \cref{fig:human}, where the CMAST framework achieves performance comparable to the human level.
We present this experiment not to claim that our method has surpassed human capabilities but to showcase its potential and the interesting phenomena observed. As for the difference manifested by the two evaluation approaches, we will discuss it in Appendix.


\section{Conclusion}
\label{sec:conclusion}
We introduce and formalize the problem of Human-centric Open-future Task Discovery, enabling LMMs to identify tasks that assist humans.
To study it, we present HOTD-Bench, a benchmark with over 2K real-world videos of diverse activities, coupled with a simulation-based protocol that evaluates open-set futures beyond observed trajectories.
For stable evaluation, we additionally provide open-vocabulary task labels annotated through a semi-automated pipeline.
We further propose the Collaborative Multi-Agent Search Tree framework, which leverages a multi-agent system and a scalable search tree module to structure the complex reasoning.
Experiments show substantial gains in Valid Task Count and Valid Task Ratio, with consistent improvements when integrated with existing LMMs.


\section{Acknowledgments}
This work was supported in part by the Fundamental Research Funds for Higher Education Institutions allocated to Sun Yat-sen University (Grants 25hytd007 and 2025RGZN009), in part by the Guangdong Provincial High-Level Young Talent Program (Grant 2025HYSPT0707), in part by the Tuoyuna Grant (HT-99982025-0564), in part by the Faculty Start-up Research Fund (Grant 67000-12255002), and in part by the Huawei Strategic Research Institute Talent Fund.

\bibliography{aaai2026}

\begin{thebibliography}{52}
\providecommand{\natexlab}[1]{#1}

\bibitem[{Aher, Arriaga, and Kalai(2023)}]{aher2023using}
Aher, G.~V.; Arriaga, R.~I.; and Kalai, A.~T. 2023.
\newblock Using large language models to simulate multiple humans and replicate human subject studies.
\newblock In \emph{International Conference on Machine Learning}, 337--371. PMLR.

\bibitem[{Ahn et~al.(2024)Ahn, Dwibedi, Finn, Arenas, Gopalakrishnan, Hausman, Ichter, Irpan, Joshi, Julian et~al.}]{ahn2024autort}
Ahn, M.; Dwibedi, D.; Finn, C.; Arenas, M.~G.; Gopalakrishnan, K.; Hausman, K.; Ichter, B.; Irpan, A.; Joshi, N.; Julian, R.; et~al. 2024.
\newblock Autort: Embodied foundation models for large scale orchestration of robotic agents.
\newblock \emph{arXiv preprint arXiv:2401.12963}.

\bibitem[{Bai et~al.(2022)Bai, Kadavath, Kundu, Askell, Kernion, Jones, Chen, Goldie, Mirhoseini, McKinnon et~al.}]{bai2022constitutional}
Bai, Y.; Kadavath, S.; Kundu, S.; Askell, A.; Kernion, J.; Jones, A.; Chen, A.; Goldie, A.; Mirhoseini, A.; McKinnon, C.; et~al. 2022.
\newblock Constitutional ai: Harmlessness from ai feedback.
\newblock \emph{arXiv preprint arXiv:2212.08073}.

\bibitem[{Bharadhwaj et~al.(2024)Bharadhwaj, Dwibedi, Gupta, Tulsiani, Doersch, Xiao, Shah, Xia, Sadigh, and Kirmani}]{bharadhwaj2024gen2act}
Bharadhwaj, H.; Dwibedi, D.; Gupta, A.; Tulsiani, S.; Doersch, C.; Xiao, T.; Shah, D.; Xia, F.; Sadigh, D.; and Kirmani, S. 2024.
\newblock Gen2Act: Human Video Generation in Novel Scenarios enables Generalizable Robot Manipulation.
\newblock \emph{arXiv preprint arXiv:2409.16283}.

\bibitem[{Caba~Heilbron et~al.(2015)Caba~Heilbron, Escorcia, Ghanem, and Carlos~Niebles}]{caba2015activitynet}
Caba~Heilbron, F.; Escorcia, V.; Ghanem, B.; and Carlos~Niebles, J. 2015.
\newblock Activitynet: A large-scale video benchmark for human activity understanding.
\newblock In \emph{Proceedings of the ieee conference on computer vision and pattern recognition}, 961--970.

\bibitem[{Cao et~al.(2024)Cao, Jia, Arik, Pfister, Zheng, Ye, and Liu}]{caotempo}
Cao, D.; Jia, F.; Arik, S.~O.; Pfister, T.; Zheng, Y.; Ye, W.; and Liu, Y. 2024.
\newblock TEMPO: Prompt-based Generative Pre-trained Transformer for Time Series Forecasting.
\newblock In \emph{International Conference on Learning Representations}.

\bibitem[{Chen et~al.(2024)Chen, Wang, Cao, Liu, Gao, Cui, Zhu, Ye, Tian, Liu et~al.}]{chen2024expanding}
Chen, Z.; Wang, W.; Cao, Y.; Liu, Y.; Gao, Z.; Cui, E.; Zhu, J.; Ye, S.; Tian, H.; Liu, Z.; et~al. 2024.
\newblock Expanding Performance Boundaries of Open-Source Multimodal Models with Model, Data, and Test-Time Scaling.
\newblock \emph{arXiv preprint arXiv:2412.05271}.

\bibitem[{Dai et~al.(2022)Dai, Das, Sharma, Minciullo, Garattoni, Bremond, and Francesca}]{dai2022toyota}
Dai, R.; Das, S.; Sharma, S.; Minciullo, L.; Garattoni, L.; Bremond, F.; and Francesca, G. 2022.
\newblock Toyota smarthome untrimmed: Real-world untrimmed videos for activity detection.
\newblock \emph{IEEE Transactions on Pattern Analysis and Machine Intelligence}, 45(2): 2533--2550.

\bibitem[{Damen et~al.(2022)Damen, Doughty, Farinella, Furnari, Kazakos, Ma, Moltisanti, Munro, Perrett, Price et~al.}]{damen2022rescaling}
Damen, D.; Doughty, H.; Farinella, G.~M.; Furnari, A.; Kazakos, E.; Ma, J.; Moltisanti, D.; Munro, J.; Perrett, T.; Price, W.; et~al. 2022.
\newblock Rescaling egocentric vision: Collection, pipeline and challenges for epic-kitchens-100.
\newblock \emph{International Journal of Computer Vision}, 1--23.

\bibitem[{Das et~al.(2019)Das, Dai, Koperski, Minciullo, Garattoni, Bremond, and Francesca}]{das2019toyota}
Das, S.; Dai, R.; Koperski, M.; Minciullo, L.; Garattoni, L.; Bremond, F.; and Francesca, G. 2019.
\newblock Toyota smarthome: Real-world activities of daily living.
\newblock In \emph{Proceedings of the IEEE/CVF international conference on computer vision}, 833--842.

\bibitem[{Driess et~al.(2023)Driess, Xia, Sajjadi, Lynch, Chowdhery, Ichter, Wahid, Tompson, Vuong, Yu et~al.}]{driess2023palm}
Driess, D.; Xia, F.; Sajjadi, M.~S.; Lynch, C.; Chowdhery, A.; Ichter, B.; Wahid, A.; Tompson, J.; Vuong, Q.; Yu, T.; et~al. 2023.
\newblock Palm-e: An embodied multimodal language model.
\newblock \emph{arXiv preprint arXiv:2303.03378}.

\bibitem[{Grauman et~al.(2022)Grauman, Westbury, Byrne, Chavis, Furnari, Girdhar, Hamburger, Jiang, Liu, Liu et~al.}]{grauman2022ego4d}
Grauman, K.; Westbury, A.; Byrne, E.; Chavis, Z.; Furnari, A.; Girdhar, R.; Hamburger, J.; Jiang, H.; Liu, M.; Liu, X.; et~al. 2022.
\newblock Ego4d: Around the world in 3,000 hours of egocentric video.
\newblock In \emph{Proceedings of the IEEE/CVF conference on computer vision and pattern recognition}, 18995--19012.

\bibitem[{Gruver et~al.(2023)Gruver, Finzi, Qiu, and Wilson}]{gruver2023large}
Gruver, N.; Finzi, M.; Qiu, S.; and Wilson, A.~G. 2023.
\newblock Large language models are zero-shot time series forecasters.
\newblock \emph{Advances in Neural Information Processing Systems}, 36: 19622--19635.

\bibitem[{Guo et~al.(2025)Guo, Yang, Zhang, Song, Zhang, Xu, Zhu, Ma, Wang, Bi et~al.}]{guo2025deepseek}
Guo, D.; Yang, D.; Zhang, H.; Song, J.; Zhang, R.; Xu, R.; Zhu, Q.; Ma, S.; Wang, P.; Bi, X.; et~al. 2025.
\newblock Deepseek-r1: Incentivizing reasoning capability in llms via reinforcement learning.
\newblock \emph{arXiv preprint arXiv:2501.12948}.

\bibitem[{Hong et~al.(2023)Hong, Zheng, Chen, Cheng, Wang, Zhang, Wang, Yau, Lin, Zhou et~al.}]{hong2023metagpt}
Hong, S.; Zheng, X.; Chen, J.; Cheng, Y.; Wang, J.; Zhang, C.; Wang, Z.; Yau, S. K.~S.; Lin, Z.; Zhou, L.; et~al. 2023.
\newblock Metagpt: Meta programming for multi-agent collaborative framework.
\newblock \emph{arXiv preprint arXiv:2308.00352}.

\bibitem[{Huang et~al.(2023)Huang, Mees, Zeng, and Burgard}]{huang2023visual}
Huang, C.; Mees, O.; Zeng, A.; and Burgard, W. 2023.
\newblock Visual language maps for robot navigation.
\newblock In \emph{2023 IEEE International Conference on Robotics and Automation (ICRA)}, 10608--10615. IEEE.

\bibitem[{Jia et~al.(2020)Jia, Chen, Huang, Zhu, and Zhu}]{jia2020lemma}
Jia, B.; Chen, Y.; Huang, S.; Zhu, Y.; and Zhu, S.-c. 2020.
\newblock LEMMA: A Multi-view Dataset for LE arning M ulti-agent M ulti-task A ctivities.
\newblock In \emph{European Conference on Computer Vision}, 767--786. Springer.

\bibitem[{Jin et~al.(2024)Jin, Wang, Ma, Chu, Zhang, Shi, Chen, Liang, Li, Pan et~al.}]{jin2024time}
Jin, M.; Wang, S.; Ma, L.; Chu, Z.; Zhang, J.; Shi, X.; Chen, P.-Y.; Liang, Y.; Li, Y.-f.; Pan, S.; et~al. 2024.
\newblock Time-LLM: Time Series Forecasting by Reprogramming Large Language Models.
\newblock In \emph{International Conference on Learning Representations}.

\bibitem[{Katara, Xian, and Fragkiadaki(2024)}]{katara2024gen2sim}
Katara, P.; Xian, Z.; and Fragkiadaki, K. 2024.
\newblock Gen2sim: Scaling up robot learning in simulation with generative models.
\newblock In \emph{2024 IEEE International Conference on Robotics and Automation (ICRA)}, 6672--6679. IEEE.

\bibitem[{Khandelwal et~al.(2022)Khandelwal, Weihs, Mottaghi, and Kembhavi}]{khandelwal2022simple}
Khandelwal, A.; Weihs, L.; Mottaghi, R.; and Kembhavi, A. 2022.
\newblock Simple but effective: Clip embeddings for embodied ai.
\newblock In \emph{Proceedings of the IEEE/CVF Conference on Computer Vision and Pattern Recognition}, 14829--14838.

\bibitem[{Lei et~al.(2018)Lei, Yu, Bansal, and Berg}]{lei2018tvqa}
Lei, J.; Yu, L.; Bansal, M.; and Berg, T.~L. 2018.
\newblock Tvqa: Localized, compositional video question answering.
\newblock \emph{arXiv preprint arXiv:1809.01696}.

\bibitem[{Liang et~al.(2023)Liang, Huang, Xia, Xu, Hausman, Ichter, Florence, and Zeng}]{liang2023code}
Liang, J.; Huang, W.; Xia, F.; Xu, P.; Hausman, K.; Ichter, B.; Florence, P.; and Zeng, A. 2023.
\newblock Code as policies: Language model programs for embodied control.
\newblock In \emph{2023 IEEE International Conference on Robotics and Automation (ICRA)}, 9493--9500. IEEE.

\bibitem[{Lin et~al.(2023{\natexlab{a}})Lin, Ye, Zhu, Cui, Ning, Jin, and Yuan}]{lin2023video}
Lin, B.; Ye, Y.; Zhu, B.; Cui, J.; Ning, M.; Jin, P.; and Yuan, L. 2023{\natexlab{a}}.
\newblock Video-llava: Learning united visual representation by alignment before projection.
\newblock \emph{arXiv preprint arXiv:2311.10122}.

\bibitem[{Lin et~al.(2023{\natexlab{b}})Lin, Agia, Migimatsu, Pavone, and Bohg}]{lin2023text2motion}
Lin, K.; Agia, C.; Migimatsu, T.; Pavone, M.; and Bohg, J. 2023{\natexlab{b}}.
\newblock Text2motion: From natural language instructions to feasible plans.
\newblock \emph{Autonomous Robots}, 47(8): 1345--1365.

\bibitem[{Moeslund, Hilton, and Kr{\"u}ger(2006)}]{moeslund2006survey}
Moeslund, T.~B.; Hilton, A.; and Kr{\"u}ger, V. 2006.
\newblock A survey of advances in vision-based human motion capture and analysis.
\newblock \emph{Computer vision and image understanding}, 104(2-3): 90--126.

\bibitem[{OpenAI(2024)}]{openaio1}
OpenAI. 2024.
\newblock Learning to Reason with LLMs.
\newblock \url{https://openai.com/index/learning-to-reason-with-llms/}.

\bibitem[{OpenAI(2025)}]{openai2025o3minisystemcard}
OpenAI. 2025.
\newblock OpenAI o3-mini System Card.
\newblock Technical report, OpenAI.
\newblock Accessed: 2025-03-04.

\bibitem[{Rafailov et~al.(2023)Rafailov, Sharma, Mitchell, Manning, Ermon, and Finn}]{rafailov2023direct}
Rafailov, R.; Sharma, A.; Mitchell, E.; Manning, C.~D.; Ermon, S.; and Finn, C. 2023.
\newblock Direct preference optimization: Your language model is secretly a reward model.
\newblock \emph{Advances in Neural Information Processing Systems}, 36: 53728--53741.

\bibitem[{Sigurdsson et~al.(2016)Sigurdsson, Varol, Wang, Farhadi, Laptev, and Gupta}]{sigurdsson2016hollywood}
Sigurdsson, G.~A.; Varol, G.; Wang, X.; Farhadi, A.; Laptev, I.; and Gupta, A. 2016.
\newblock Hollywood in homes: Crowdsourcing data collection for activity understanding.
\newblock In \emph{Computer Vision--ECCV 2016: 14th European Conference, Amsterdam, The Netherlands, October 11--14, 2016, Proceedings, Part I 14}, 510--526. Springer.

\bibitem[{Soomro(2012)}]{soomro2012ucf101}
Soomro, K. 2012.
\newblock UCF101: A dataset of 101 human actions classes from videos in the wild.
\newblock \emph{arXiv preprint arXiv:1212.0402}.

\bibitem[{Team(2024)}]{internvl2}
Team, O. 2024.
\newblock InternVL2: Better than the Best—Expanding Performance Boundaries of Open-Source Multimodal Models with the Progressive Scaling Strategy.
\newblock \url{https://internvl.github.io/blog/2024-07-02-InternVL-2.0/}.

\bibitem[{Team(2025)}]{qwen2.5-VL}
Team, Q. 2025.
\newblock Qwen2.5-VL.

\bibitem[{Wang et~al.(2023{\natexlab{a}})Wang, Ling, Yuan, Shridhar, Bao, Qin, Wang, Xu, and Wang}]{wang2023gensim}
Wang, L.; Ling, Y.; Yuan, Z.; Shridhar, M.; Bao, C.; Qin, Y.; Wang, B.; Xu, H.; and Wang, X. 2023{\natexlab{a}}.
\newblock Gensim: Generating robotic simulation tasks via large language models.
\newblock \emph{arXiv preprint arXiv:2310.01361}.

\bibitem[{Wang et~al.(2024)Wang, Bai, Tan, Wang, Fan, Bai, Chen, Liu, Wang, Ge et~al.}]{wang2024qwen2}
Wang, P.; Bai, S.; Tan, S.; Wang, S.; Fan, Z.; Bai, J.; Chen, K.; Liu, X.; Wang, J.; Ge, W.; et~al. 2024.
\newblock Qwen2-vl: Enhancing vision-language model's perception of the world at any resolution.
\newblock \emph{arXiv preprint arXiv:2409.12191}.

\bibitem[{Wang et~al.(2023{\natexlab{b}})Wang, Xian, Chen, Wang, Wang, Fragkiadaki, Erickson, Held, and Gan}]{wang2023robogen}
Wang, Y.; Xian, Z.; Chen, F.; Wang, T.-H.; Wang, Y.; Fragkiadaki, K.; Erickson, Z.; Held, D.; and Gan, C. 2023{\natexlab{b}}.
\newblock Robogen: Towards unleashing infinite data for automated robot learning via generative simulation.
\newblock \emph{arXiv preprint arXiv:2311.01455}.

\bibitem[{Wu et~al.(2023)Wu, Bansal, Zhang, Wu, Zhang, Zhu, Li, Jiang, Zhang, and Wang}]{wu2023autogen}
Wu, Q.; Bansal, G.; Zhang, J.; Wu, Y.; Zhang, S.; Zhu, E.; Li, B.; Jiang, L.; Zhang, X.; and Wang, C. 2023.
\newblock Autogen: Enabling next-gen llm applications via multi-agent conversation framework.
\newblock \emph{arXiv preprint arXiv:2308.08155}.

\bibitem[{Yang et~al.(2024{\natexlab{a}})Yang, Yang, Hui, Zheng, Yu, Zhou, Li, Li, Liu, Huang, Dong, Wei, Lin, Tang, Wang, Yang, Tu, Zhang, Ma, Yang, Xu, Zhou, Bai, He, Lin, Dang, Lu, Chen, Yang, Li, Xue, Ni, Zhang, Wang, Peng, Men, Gao, Lin, Wang, Bai, Tan, Zhu, Li, Liu, Ge, Deng, Zhou, Ren, Zhang, Wei, Ren, Liu, Fan, Yao, Zhang, Wan, Chu, Liu, Cui, Zhang, Guo, and Fan}]{yang2024qwen2technicalreport}
Yang, A.; Yang, B.; Hui, B.; Zheng, B.; Yu, B.; Zhou, C.; Li, C.; Li, C.; Liu, D.; Huang, F.; Dong, G.; Wei, H.; Lin, H.; Tang, J.; Wang, J.; Yang, J.; Tu, J.; Zhang, J.; Ma, J.; Yang, J.; Xu, J.; Zhou, J.; Bai, J.; He, J.; Lin, J.; Dang, K.; Lu, K.; Chen, K.; Yang, K.; Li, M.; Xue, M.; Ni, N.; Zhang, P.; Wang, P.; Peng, R.; Men, R.; Gao, R.; Lin, R.; Wang, S.; Bai, S.; Tan, S.; Zhu, T.; Li, T.; Liu, T.; Ge, W.; Deng, X.; Zhou, X.; Ren, X.; Zhang, X.; Wei, X.; Ren, X.; Liu, X.; Fan, Y.; Yao, Y.; Zhang, Y.; Wan, Y.; Chu, Y.; Liu, Y.; Cui, Z.; Zhang, Z.; Guo, Z.; and Fan, Z. 2024{\natexlab{a}}.
\newblock Qwen2 Technical Report.
\newblock arXiv:2407.10671.

\bibitem[{Yang et~al.(2024{\natexlab{b}})Yang, Luo, Pani, and Yang}]{yang2024bbsea}
Yang, S.; Luo, Q.; Pani, A.; and Yang, Y. 2024{\natexlab{b}}.
\newblock BBSEA: An Exploration of Brain-Body Synchronization for Embodied Agents.
\newblock \emph{arXiv preprint arXiv:2402.08212}.

\bibitem[{Yang et~al.(2024{\natexlab{c}})Yang, Zhou, Li, Tao, Li, Shen, He, Jiang, and Shi}]{yang2024embodied}
Yang, Y.; Zhou, T.; Li, K.; Tao, D.; Li, L.; Shen, L.; He, X.; Jiang, J.; and Shi, Y. 2024{\natexlab{c}}.
\newblock Embodied multi-modal agent trained by an llm from a parallel textworld.
\newblock In \emph{Proceedings of the IEEE/CVF Conference on Computer Vision and Pattern Recognition}, 26275--26285.

\bibitem[{Yang et~al.(2024{\natexlab{d}})Yang, Liu, Chen, Cherian, Marks, Le~Roux, and Gan}]{yang2024rila}
Yang, Z.; Liu, J.; Chen, P.; Cherian, A.; Marks, T.~K.; Le~Roux, J.; and Gan, C. 2024{\natexlab{d}}.
\newblock RILA: Reflective and Imaginative Language Agent for Zero-Shot Semantic Audio-Visual Navigation.
\newblock In \emph{Proceedings of the IEEE/CVF Conference on Computer Vision and Pattern Recognition}, 16251--16261.

\bibitem[{Yu et~al.(2023)Yu, Gileadi, Fu, Kirmani, Lee, Arenas, Chiang, Erez, Hasenclever, Humplik et~al.}]{yu2023language}
Yu, W.; Gileadi, N.; Fu, C.; Kirmani, S.; Lee, K.-H.; Arenas, M.~G.; Chiang, H.-T.~L.; Erez, T.; Hasenclever, L.; Humplik, J.; et~al. 2023.
\newblock Language to rewards for robotic skill synthesis.
\newblock \emph{arXiv preprint arXiv:2306.08647}.

\bibitem[{Yu et~al.(2025)Yu, Yao, Li, Deng, Jiang, Cao, Chen, Suchow, Cui, Liu et~al.}]{yu2025fincon}
Yu, Y.; Yao, Z.; Li, H.; Deng, Z.; Jiang, Y.; Cao, Y.; Chen, Z.; Suchow, J.; Cui, Z.; Liu, R.; et~al. 2025.
\newblock Fincon: A synthesized llm multi-agent system with conceptual verbal reinforcement for enhanced financial decision making.
\newblock \emph{Advances in Neural Information Processing Systems}, 37: 137010--137045.

\bibitem[{Yuan et~al.(2024)Yuan, Cao, Li, Jiang, and Wang}]{yuan2024sd}
Yuan, Z.; Cao, J.; Li, Z.; Jiang, H.; and Wang, Z. 2024.
\newblock Sd-mvs: Segmentation-driven deformation multi-view stereo with spherical refinement and em optimization.
\newblock In \emph{Proceedings of the AAAI conference on artificial intelligence}, volume~38, 6871--6880.

\bibitem[{Yuan et~al.(2025{\natexlab{a}})Yuan, Qu, Qian, Chen, Tang, Sun, Chu, Zhang, Wang, Cai et~al.}]{yuan2025video}
Yuan, Z.; Qu, X.; Qian, C.; Chen, R.; Tang, J.; Sun, L.; Chu, X.; Zhang, D.; Wang, Y.; Cai, Y.; et~al. 2025{\natexlab{a}}.
\newblock Video-star: Reinforcing open-vocabulary action recognition with tools.
\newblock \emph{arXiv preprint arXiv:2510.08480}.

\bibitem[{Yuan et~al.(2025{\natexlab{b}})Yuan, Tang, Luo, Chen, Qian, Sun, Chu, Cai, Zhang, and Li}]{yuan2025autodrive}
Yuan, Z.; Tang, J.; Luo, J.; Chen, R.; Qian, C.; Sun, L.; Chu, X.; Cai, Y.; Zhang, D.; and Li, S. 2025{\natexlab{b}}.
\newblock AutoDrive-R2: Incentivizing Reasoning and Self-Reflection Capacity for VLA Model in Autonomous Driving.
\newblock \emph{arXiv preprint arXiv:2509.01944}.

\bibitem[{Yuan et~al.(2025{\natexlab{c}})Yuan, Yang, Cai, Wu, Liu, Zhang, Jiang, Li, and Wang}]{yuan2025sed}
Yuan, Z.; Yang, Z.; Cai, Y.; Wu, K.; Liu, M.; Zhang, D.; Jiang, H.; Li, Z.; and Wang, Z. 2025{\natexlab{c}}.
\newblock SED-MVS: Segmentation-Driven and Edge-Aligned Deformation Multi-View Stereo with Depth Restoration and Occlusion Constraint.
\newblock \emph{IEEE Transactions on Circuits and Systems for Video Technology}.

\bibitem[{Zellers et~al.(2019)Zellers, Bisk, Farhadi, and Choi}]{zellers2019recognition}
Zellers, R.; Bisk, Y.; Farhadi, A.; and Choi, Y. 2019.
\newblock From recognition to cognition: Visual commonsense reasoning.
\newblock In \emph{Proceedings of the IEEE/CVF conference on computer vision and pattern recognition}, 6720--6731.

\bibitem[{Zeng et~al.(2024)Zeng, Chen, Liang, Wu, Cao, and Guo}]{zeng2024benchmarking}
Zeng, R.; Chen, X.; Liang, J.; Wu, H.; Cao, G.; and Guo, Y. 2024.
\newblock Benchmarking the Robustness of Temporal Action Detection Models Against Temporal Corruptions.
\newblock In \emph{Proceedings of the IEEE/CVF Conference on Computer Vision and Pattern Recognition}, 18263--18274.

\bibitem[{Zhang et~al.(2024)Zhang, Li, Liu, Lee, Gui, Fu, Feng, Liu, and Li}]{zhang2024llavanextvideo}
Zhang, Y.; Li, B.; Liu, h.; Lee, Y.~j.; Gui, L.; Fu, D.; Feng, J.; Liu, Z.; and Li, C. 2024.
\newblock LLaVA-NeXT: A Strong Zero-shot Video Understanding Model.

\bibitem[{Zhao, Weber, and Wermter(2024)}]{zhao2024agentic}
Zhao, X.; Weber, C.; and Wermter, S. 2024.
\newblock Agentic Skill Discovery.
\newblock \emph{arXiv preprint arXiv:2405.15019}.

\bibitem[{Zhou et~al.(2023)Zhou, Mart{\'\i}n-Mart{\'\i}n, Kapadia, Savarese, and Niebles}]{zhou2023procedure}
Zhou, H.; Mart{\'\i}n-Mart{\'\i}n, R.; Kapadia, M.; Savarese, S.; and Niebles, J.~C. 2023.
\newblock Procedure-aware pretraining for instructional video understanding.
\newblock In \emph{Proceedings of the IEEE/CVF Conference on Computer Vision and Pattern Recognition}, 10727--10738.

\bibitem[{Zhou et~al.(2024)Zhou, Atreya, Lee, Walke, Mees, and Levine}]{zhou2024autonomous}
Zhou, Z.; Atreya, P.; Lee, A.; Walke, H.; Mees, O.; and Levine, S. 2024.
\newblock Autonomous improvement of instruction following skills via foundation models.
\newblock \emph{arXiv preprint arXiv:2407.20635}.

\end{thebibliography}

\end{document}